\documentclass{article}

\usepackage{spconf}
\usepackage{balance}

\name{
  Jan Funke$^{\star\ddagger}$ \qquad %
  Francesc Moreno-Noguer$^\star$ \qquad %
  Albert Cardona$^\dagger$ \qquad %
  Matthew Cook$^\ddagger$}
\address
{
\begin{minipage}{0.3\textwidth}
  \centering
  $^{\star}$ Institut de Rob\`otica i Inform\`atica Industrial\\UPC/CSIC Barcelona
\end{minipage}
\begin{minipage}{0.3\textwidth}
  \centering
  $^{\dagger}$ Janelia Research Campus\\VA, Ashburn\\
\end{minipage}
\begin{minipage}{0.3\textwidth}
  \centering
  $^{\ddagger}$ Institute of Neuroinformatics\\UZH/ETH Zurich
\end{minipage}
}
\usepackage{amsmath}
\usepackage{amssymb}
\usepackage{dsfont}
\usepackage{tikz}
\usepackage{xifthen}
\usepackage{caption}
\usepackage{subcaption}
\usepackage{funkey}
\usepackage[ruled,vlined]{algorithm2e}

\usetikzlibrary{calc}
\usetikzlibrary{positioning}

\usetikzlibrary{arrows}
\usetikzlibrary{backgrounds}
\usetikzlibrary{calc}
\usetikzlibrary{chains}
\usetikzlibrary{fit}
\usetikzlibrary{matrix}
\usetikzlibrary{mindmap}
\usetikzlibrary{pgfplots.colormaps}
\usetikzlibrary{positioning}
\usetikzlibrary{shadows}
\usetikzlibrary{shapes}
\usetikzlibrary{trees}

\tikzstyle{inactive}=
  [draw=gray,text=gray]

\tikzstyle{data}=
  [rectangle,rounded corners,draw,fill=gray!10!white]
\tikzstyle{method}=
  [rounded rectangle,draw]

\tikzstyle{image}=
  [rectangle,draw,fill=white,inner sep=2pt]

\tikzstyle{process}=[
  single arrow,
  single arrow head extend=0.125cm,
  inner sep=1pt,
  minimum height=0.5cm,
  fill=gray]

\tikzstyle{slice}=
  [rectangle,draw=black,inner sep=10pt]

\tikzstyle{candidate}=
  [circle,fill=white,draw=black]

\tikzstyle{assignment}=
  [circle,fill=assignmentcolor,draw=assignmentcolor!75!black,inner sep=1pt]

\tikzstyle{connection}=
  [draw=black,thick]

\tikzstyle{subset}=
  [<-,draw=black]

\tikzstyle{feature}=
  [anchor=base]

\tikzstyle{var}=
  [circle,fill=white,draw=black,inner sep=3pt]

\tikzstyle{factor}=
  [rectangle,fill=gray,draw=black,inner sep=5pt]

\tikzstyle{region}=
  [rounded rectangle,opacity=0.5]

\tikzstyle{dependency}=
  [<-,>=latex',thick]

\tikzstyle{task}=
  [rectangle,rounded corners=0.1cm,draw,fill=orange!50!white,inner sep=5pt]

\def\addlegendimage{\csname pgfplots@addlegendimage\endcsname}
\def\addlegendentry{\csname pgfplots@addlegendentry\endcsname}

\pgfplotsset{compat=1.5}
\usetikzlibrary{pgfplots.groupplots}

\pgfplotsset{
  errors/.style={
    stack plots=y,
    area style,
    enlarge x limits=false,
    xmajorgrids=true,
    ymajorgrids=true,
    yminorgrids=true,
    legend reversed
  }
}

\usepackage{xspace}

\def\reffig#1{Figure~\ref{#1}}

\def\refeq#1{(\ref{#1})}

\makeatletter
\def\etal{et al\@ifnextchar.{}{.\@\xspace}}
\makeatother

\def\sopnet{{\normalfont\textsc{Sop\-net}}\xspace}

\def\ted{TED\xspace}

\DeclareMathOperator\TEDsplit{s}
\DeclareMathOperator\TEDmerge{m}
\DeclareMathOperator\TED{TED}

\pgfplotsset{every axis/.append style={small,legend style={font=\small}}}

\usepackage[pagebackref=true,breaklinks=true,letterpaper=true,colorlinks,bookmarks=false]{hyperref}

\def\StripPrefix#1>{}
\def\jobis#1{FF\fi
  \def\predicate{#1}%
  \edef\predicate{\expandafter\StripPrefix\meaning\predicate}%
  \edef\job{\jobname}%
  \ifx\job\predicate
}

\title{TED: A Tolerant Edit Distance for Segmentation Evaluation}%

\begin{document}

\maketitle
\thispagestyle{empty}

\begin{abstract}

%
In this paper, we present a novel error measure to compare a segmentation against ground truth.
This measure, which we call Tolerant Edit Distance (\ted), is motivated by two observations:
(1) Some errors, like small boundary shifts, are tolerable in practice. Which errors are tolerable is application dependent and should be a parameter of the measure.
%
(2)~Non-tolerable errors have to be corrected manually. The time needed to do so should be reflected by the error measure.
Using integer linear programming, the \ted finds the minimal weighted sum of split and merge errors exceeding a given tolerance criterion, and thus provides a time-to-fix estimate. 
In contrast to commonly used measures like Rand index or variation of information, the \ted (1) does not count small, but tolerable, differences, (2) provides intuitive numbers, (3) gives a time-to-fix estimate, and (4) can localize and classify the type of errors.
By supporting both isotropic and anisotropic volumes and having a flexible tolerance criterion, the \ted can be adapted to different requirements.
On example segmentations for 3D neuron segmentation, we demonstrate that the \ted is capable of counting topological errors, while ignoring small boundary shifts.

\end{abstract}

\section{Introduction}
\label{sec:introduction}

%
In the computer vision literature, several approaches to assess the quality of
contour detection and segmentation algorithms can be found.
%
  Most of these measures have been designed to capture the intuition of what
  humans consider to be two similar results. In particular, these measures are
  supposed to be robust to certain tolerated deviations, like small shifts of
  contours.
  %
  For the contour detection in the Berkeley segmentation
  dataset~\cite{Martin2001}, for example, the precision and recall of detected
  boundary pixels within a threshold distance to the ground truth became the
  widely used standard~\cite{Martin2004,Arbelaez2009}.
  %
  Contour error measures are, however, not a good fit for segmentations, since
  small errors in the detection of a contour can lead to the split or merge of
  segments. Therefore, alternatives like the Variation Of Information
  (VOI), the Rand Index~\cite{Rand1971} (RI), the probabilistic Rand
  index~\cite{Unnikrishnan2005,Unnikrishnan2007}, and the segmentation covering
  measure~\cite{Arbelaez2009}, have been proposed.

\noindent
\begin{minipage}{0.4835\textwidth}
  \centerline{\includetikz[scale=0.95]{figures/ted/ted_example}}
  \vspace{-3mm}
  
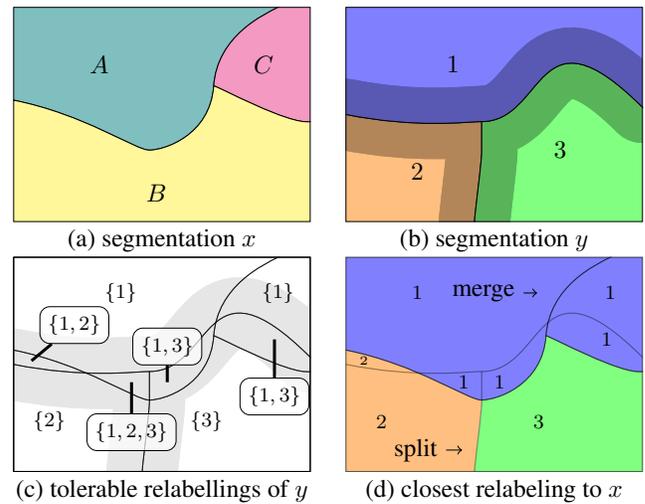
\captionof{figure}{\label{fig:ted:illustration}Illustration of the Tolerant Edit Distance (TED) between two  segmentations $x$ and $y$. By tolerating boundary shifts to a certain extend, shown as shadow in (b), $y$ is allowed to be changed to match $x$ as closely as possible. For that, we consider regions obtained by combining the $x$ and $y$, illustrated in (c). For each of these regions, we enumerate a set of labels used by $y$ that are within a threshold distance to all locations inside the region (shown in curly brackets). This threshold is the maximally allowed boundary shift. Note that in this example, the region obtained from intersecting $A$ and $3$ can change its label to $1$ (or keep $3$), but not to $2$, since it contains points that are too far way from region $2$. Regions with only one possible label are too large to be relabeled by shifting their boundary and have to keep their initial label. From all the possible ways to relabel $y$, the relabeling (d) minimizing the number of split and merge errors compared to $x$ is chosen by solving an integer linear program.}
\end{minipage}
\vspace{2mm}

%
However, these measures do not acknowledge that there are different criteria for segmentation comparison, and instead accumulate errors uniformly, even for many small differences that are irrelevant in practice.
  Especially in the field of biomedical image processing, we are often more interested in counting true topological errors like splits and merges of objects, instead of counting small deviations from the ground truth contours. This is in particular the case for imaging methods for which no unique ``ground truth'' labeling exists. In the imaging of neural tissue with Electron Microscopy (EM), for example, the preparation protocol can alter the volume of neural processes, such that it is hard to know what the true size was~\cite{Sosinsky2008}. Further, the imaging resolution and data quality might just not be sufficient to clearly locate contours between objects~\cite{Cardona2012a}, resulting in a high inter-observer variability.

%
To address these issues, we present a novel measure to evaluate segmentations on a clearly specified tolerance criterion:
  At the core of our measure, that we call \emph{Tolerant Edit Distance} (\ted)\footnote{Source code available at \url{http://github.com/funkey/ted}.}, is an explicit tolerance criterion (\eg, boundary shifts within a certain range). Using integer linear programming, we find the minimal weighted sum of split and merge errors exceeding the tolerance criterion, and thus provide a time-to-fix estimate. By interpreting a segmentation as a general labeling of voxels, our measure does not require voxels of the same object to form a connected component, and thus supports anisotropic volumes, missing data, or known object connections via paths outside the volume being considered. The reported results are intuitive, easy to interpret, and errors can be localized in the volume. An illustration of the \ted can be found in \reffig{fig:ted:illustration}.

\begin{figure}[t]
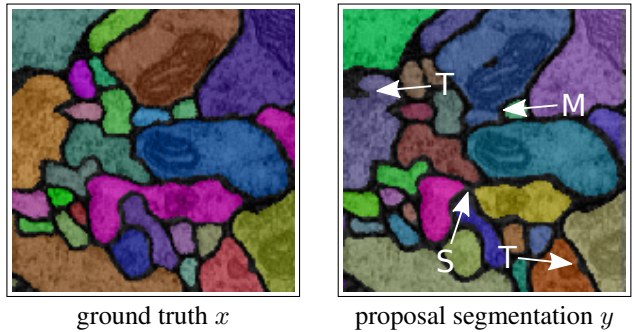

  \centerline{%
    \includetikz{figures/introduction/small_errors}%
  }
  \vspace{-0.5cm}
  \caption{Example errors made by an automatic neuron segmentation algorithm. Errors like merges (M) and splits (S) dramatically change the reconstructed topology and should be avoided. Small disagreements in the boundary location (T) are however tolerable and should be ignored during evaluation.}
  \label{fig:introduction:small_errors}
  \vspace{-0.25cm}
\end{figure}

\vspace{1mm}
\noindent
{\bf Application to Neuron Segmentation.}
%
%
To demonstrate the usefulness of our measure, we present our results in the context of automatic neuron segmentation from EM volumes, an active field of biomedical image processing (for recent advances, see~\cite{Funke2012,Kaynig2013,Nunez-Iglesias2013,Parag2014,Huang2014}).
  In this field, the criterion to assess the quality of a segmentation depends on the biological question:
  On one hand, \emph{skeletons} of neurons are sufficient to identify individual neurons~\cite{Peng2011}, to study neuron types and their function~\cite{Denk2012}, and to obtain the wiring diagram of a nervous system (the so-called \emph{connectome})~\cite{Cardona2012a}. In these cases, topological correctness is far more important than the diameter of a neural process or the exact location of its boundary (see \reffig{fig:introduction:small_errors} for examples).
  On the other hand, for biophysically realistic neuron simulation, \emph{volumetric} information is needed to model action potential time dynamics, and to understand and simulate information processing capabilities of single neurons~\cite{London2005}. In this case, the segmentation should be close to the true volume of the reconstructed neurons. Only small deviations in the boundary location might still be tolerable.

%
%

Current state-of-the-art methods for automatic neuron segmentation can broadly be divided into isotropic~\cite{Nunez-Iglesias2013,Kroeger2013,Parag2014,Huang2014} and anisotropic methods~\cite{Funke2012,Kaynig2013,Funke2014}.
  For both types, reporting segmentation accuracy in terms of VOI or RI  became the de-facto standard~\cite{Nunez-Iglesias2013,Kroeger2013,Kaynig2013,Parag2014,Huang2014}. Less frequently used~\cite{Funke2012,Funke2014} is the \emph{Anisotropic Edit Distance} (AED)~\cite{Funke2012} and the \emph{Warping Error} (WE)~\cite{Jain2010b}. The AED is tailored to the specific error correction steps required for anisotropic volumes (splits and merges of 2D neuron slices within a section, connections and disconnections of slices between sections). The WE aims to measure the difference between ground truth and a proposal segmentation in terms of their topological differences. As such, the WE was the first error measure for neuron segmentation that deals with the delicate question of up to which point a boundary shift is not considered to be an error. However, since the WE assumes a foreground-background segmentation where connected foreground objects represent neurons, it is only applicable to isotropic volumes (in anisotropic volumes, connectedness of neurons is not always preserved). Furthermore, only suboptimal solutions to the WE are found using a greedy, randomized heuristic, which makes it difficult to use for evaluation purposes. Consequently, the WE has found its main application in the training of neural networks for image classification~\cite{Jain2010b}.

\section{Tolerant Edit Distance}
\label{sec:ted}

The TED measures the difference between two segmentations $x : \Omega \mapsto K_x$ and $y: \Omega \mapsto K_y$, where $\Omega$ is a discrete set of voxel (or supervoxel) locations in a volume, and $K_x$ and $K_y$ are sets of labels used by $x$ and $y$, respectively. The difference is reported in terms of the minimal number of splits and merges appearing in a relabeling of $y$, as compared with $x$. How $y$ is allowed to be relabeled is defined on a tolerance criterion, \eg, the maximal displacement of an object boundary.

We say that a label $k \in K_x$ overlaps with a label $l \in K_y$, if there exists at least one location $i \in \Omega$ such that $x(i)=k$ and $y(i)=l$. If $x$ and $y$ represent the same segmentation, each label $l$ overlaps with exactly one label $k$, and vice versa. Consequently, if a label $k \in K_x$ overlaps with $n$ labels from $K_y$, we count it as $n-1$ splits. Analogously, if a label $l \in K_y$ overlaps with $n$ labels from $K_x$, we count it as $n-1$ merges. For two labelings $x$ and $y$, we denote as $\TEDsplit(x,y)$ and $\TEDmerge(x,y)$ the sum of splits and merges over all labels.

Let a tolerance function $T$ be a binary indicator on two labeling functions $y$ and $y'$,
\begin{equation}
  T(y, y') = \left\{
  \begin{array}{ll}
    1 & \text{if $y$ is similar to $y'$,} \\
    0 & \text{otherwise.}
  \end{array}
  \right.
\end{equation}
Further, let $\mathcal{Y}$ be the set of all labeling functions $y': \Omega 
\mapsto K_y$, \ie, all possible labelings of $\Omega$ using the labels of $y$, 
and let $\mathcal{Y}^+(y) = \{ y' \in \mathcal{Y} \;|\; T(y,y') = 1 \}$ be the 
set of all tolerated relabelings of $y$.
The TED is the minimal weighted sum of splits and merges over all tolerable 
relabelings $\mathcal{Y}^+(y)$:
\begin{equation}
  \TED(x,y)
  =
  \min_{y' \in \mathcal{Y}^+(y)}
    \alpha\TEDsplit(x,y') +
    \beta\TEDmerge(x,y')
  \text{,}
  \label{eq:ted}
\end{equation}
where the weights $\alpha$ and $\beta$ represent the time or effort needed to fix a split or merge, respectively.

In order to find the minimum of \refeq{eq:ted}, we assume that the tolerance function is \emph{local}, \ie, there exists a set $A_i$ of tolerable labels for each location $i$, and a tolerable labeling is any combination of those labels:
$$
  T(y,y')
  =
  \prod_{i \in \Omega}
  \mathds{1}\left(y'(i) \in A_i\right)
  \text{.}
$$
An example of such a tolerance function is shown in \reffig{fig:ted:illustration}~(c). With this assumption, we solve \refeq{eq:ted} with the following integer linear program (ILP):
\beginilp
  \objective{\min_{\vct{v}}}{\hspace{2cm}\alpha s + \beta m}
  \constraint
    {\sum_{l \in A_i} v_{i\leftarrow l}}
    {=}
    {1}
    {\forall i \in \Omega}
    \label{eq:ted:constraints:v:1}
  \constraint
    {\sum_{i \in \Omega} v_{i\leftarrow l}}
    {\ge}
    {1}
    {\forall l \in K_y}
    \label{eq:ted:constraints:v:2}
  \constraint
    {a_{kl} - v_{i\leftarrow l}}
    {\geq}
    {0}
    {\forall i \in \Omega : x(i) = k}
    \label{eq:ted:constraints:a:1}
  \constraint
    {\hspace{-1cm}a_{kl} - \sum_{i \in \Omega :x(i) = k} v_{i\leftarrow l}}
    {\leq}
    {0}
    {\forall k\in K_x\;\;\forall l\in K_y}
    \label{eq:ted:constraints:a:2}
  \constraint
    {s_k - \sum_{l \in K_y} a_{kl}}
    {=}
    {-1}
    {\forall k\in K_x}
    \label{eq:ted:constraints:sk}
  \constraint
    {m_l - \sum_{k \in K_x} a_{kl}}
    {=}
    {-1}
    {\forall l\in K_y}
    \label{eq:ted:constraints:ml}
  \constraint
    {s - \sum_{k \in K_x} s_k}
    {=}
    {0}
    {}
    \label{eq:ted:constraints:s}
  \constraint
    {m - \sum_{l \in K_y} m_l}
    {=}
    {0}
    {}
    \label{eq:ted:constraints:m}
\endilp
At the core of this ILP are binary indicator variables $\vct{v} = (v_{i\leftarrow l} \in \{0,1\}\;|\;i \in \Omega,\;l \in A_i)$ to indicate the assignment of label $l$ to location $i$. Constraints \refeq{eq:ted:constraints:v:1} and \refeq{eq:ted:constraints:v:2} ensure that exactly one of the labels gets chosen for each location and that each label of $y$ has to appear at least once.
Further, we introduce binary variables $a_{kl}$ that indicate the presence of a joint assignment of label $k$ from $x$ and label $l$ from $y'$ at at least one location. With constraints \refeq{eq:ted:constraints:a:1} and \refeq{eq:ted:constraints:a:2} we make sure that each $a_{kl} = 1$ if and only if there is at least one location $i \in \Omega$ such that $x(i) = k$ and $y'(i) = l$.
To count the number of times a label $k\in K_x$ is split in $y'$, we further introduce integers $s_k \in \mathbb{N}$. These counts equal the number of times $k$ was matched with any other label minus one, which we ensure with constraints \refeq{eq:ted:constraints:sk}. Analogously, we introduce integers $m_l$ and constraints \refeq{eq:ted:constraints:ml} for merges caused by label $l$ in $y'$.
The final split and merge numbers $s$ and $m$ are just the sums of the label-wise splits and merges, ensured by \refeq{eq:ted:constraints:s} and \refeq{eq:ted:constraints:m}.

Once the optimal solution of this ILP has been found, the variables $a_{kl}$ can be used to determine which labels got split and merged, and thus to localize errors.

\section{Results}
\label{sec:results}

\def\snemidataset{\textsc{Mouse Cortex}\xspace}
\def\drosodataset{\textsc{Drosophila}\xspace}

\begin{figure*}
  \centering
  \begin{subfigure}[b]{0.3\linewidth}
    \includegraphics{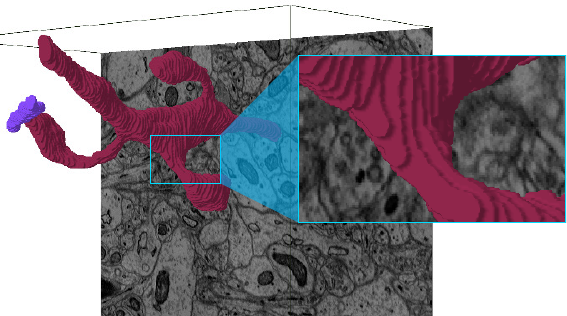}
    \caption{ground truth $x$}
  \end{subfigure}
  \hspace{5mm}
  \begin{subfigure}[b]{0.3\linewidth}
    \includegraphics{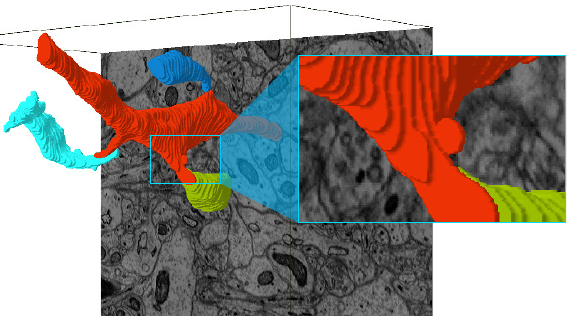}
    \caption{proposal segmentation $y$}
  \end{subfigure}
  \hspace{5mm}
  \begin{subfigure}[b]{0.3\linewidth}
      \includegraphics{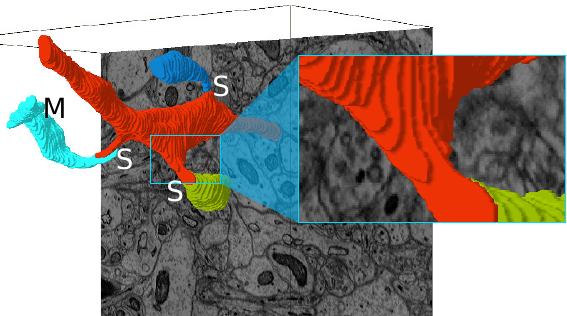}
    \caption{closest tolerable relabeling of $y$ to $x$}
  \end{subfigure}
  \caption{Errors found by the \ted between a human generated ground truth $x$ (a) and a proposal segmentation $y$ (b), illustrated on two neurons (purple and red in ground truth). Small errors, as the one shown in the magnification, are tolerated and consequently removed in the tolerable relabelling of $y$ (c). Remaining errors are considered real splits (S) and merges (M).}
  \label{fig:results:example}
\end{figure*}

\noindent
{\bf Shift of Object Boundary.}
To illustrate the behaviour of different error measures in the case of object boundary displacements, we created a simple artificial 1D labeling consisting of two regions. We show the errors of segmentations obtained by shifting the boundary between the objects.
\begin{figure}[h]
  \centerline{
    \includetikz[scale=0.7]{figures/results/comparison/boundary_shift}
  }
  \vspace{-3mm}
\end{figure}

It can clearly be seen that \ted assigns the same numbers (one split and one merge error) as soon as a given tolerance criterion is exceeded ($0.025$ in this example), regardless where the error happens. This is the desired outcome for applications like neuron segmentation, where it is important to count the number of topological errors regardless of how many voxels got affected.

\vspace{1mm}
\noindent
{\bf Influence of Distance Threshold.}
In order to study the effect of the threshold distance for boundary shifts, we used an automatic segmentation result\footnote{Obtained using \sopnet~\cite{Funke2012} on a publicly available EM dataset~\cite{Gerhard2013}} and evaluated the \ted for varying thresholds.
\begin{figure}[h]
  \def\plotwidth{11cm}
  \def\plotheight{6cm}
  \centerline{\includetikz[scale=0.7]{figures/results/tolerance/tolerance}}
  \vspace{-3mm}
\end{figure}

The \ted reveals that most of the errors occur within the range of about $50nm$, corresponding to about 12 pixels in the x-y-plane of this dataset. Depending on the biological need, those errors might be tolerable. In the same plot, we show the VOI of the closest tolerable relabeling to the ground truth under the given boundary shift threshold (\ie, the equivalent of \reffig{fig:ted:illustration}~(d) on the proposal segmentation). From this example, we can see that the errors $<50nm$ contribute quite significantly with $0.23$ bits to the total VOI of $0.886$, and thus can shadow true topological errors.

\vspace{1mm}
\noindent
{\bf Comparison to RI and VOI.}
We compare RI and VOI against \ted for three manual modifications of the ground truth labeling of~\cite{Gerhard2013}.
\begin{figure}[h]
\def\totalwidth{1024}%
\def\totalheight{1024}%
\def\samplewidth{2.5cm}%
\def\sampleheight{2.5cm}%
\vspace{-6mm}
\begin{tikzpicture}%

  \node[image] (o) at (0,0.25) {
      \includesubimage{500}{500}{200}{200}{figures/results/grow/08_original}};%
  \node[image] (g) at (2.25,-0.25) {
      \includesubimage{500}{500}{200}{200}{figures/results/grow/08_grown}};%

  \node[below of=o,yshift=-0.6cm] {original};
  \node[below of=g,yshift=-0.6cm] {shifted by $10nm$};

  \node at (5.5,-0.3) {%
    \def\plotwidth{4.5cm}%
    \def\plotheight{2.5cm}%
    \includetikz{figures/results/comparison/modifications}%
  };
\end{tikzpicture}%
\vspace{-8mm}
\end{figure}

%
For the \emph{$10nm$ shift} experiment, we shifted the boundaries of neurons in the ground truth by $10nm$. For the \emph{splits} and \emph{merges} experiment, we split and merged neurons at 10 randomly selected locations, respectively. It can be seen that the small shifts of object boundaries can have a significant contribution to the measures RI and VOI, which confirms our previous observation.

\vspace{1mm}
\noindent
{\bf Localization of Errors.}
Due to the explicit tolerance criterion of the \ted, errors can be localized in the volume. In \reffig{fig:results:example} we show example split an merge errors detected by the \ted on an automatic segmentation result for the SNEMI dataset~\cite{Arganda-Carreras2013}. The boundary shift tolerance was set to $100nm$, which corresponds to $16.6\times16.6\times3.3$ voxels for this volume with a resolution of $6nm\times 6nm\times 30nm$.

\section{Conclusions}

We presented the \ted, a novel measure for segmentation comparison, which tolerates small errors based on an explicit tolerance criterion.

Although we demonstrated the \ted in the domain of neuron segmentation, our error measure is not intrinsically limited to this application. In our future work, we will investigate its use for other computer vision problems, and especially on the training of algorithms to minimize this error measure.

A current limitation of the \ted is the restriction to use local tolerance
functions. Although more involved tolerance criteria could in theory be
incorporated into the ILP by adding auxiliary variables, it remains
questionable whether the resulting problem is still tractable. Although we did
not observe that empirically, even with the current formulation it is
conceivable that an optimal solution to the ILP can not be found in reasonable
time. This could in particular be the case if ground truth and proposal segmentation
differ a lot and a very lax tolerance criterion is used. In these cases,
approximate solutions to the proposed ILP might be considered.

\balance

{
\bibliographystyle{ieee}
\bibliography{bibliography}
}

\end{document}